\documentclass{article}
\usepackage{times}
\usepackage{amssymb}
\usepackage{amsmath}
\usepackage{bm}
\usepackage{graphicx}
\usepackage{subcaption} 
\usepackage{natbib}
\usepackage{algorithm}
\usepackage{algorithmic}
\usepackage{xspace}
\usepackage{hyperref}
\usepackage{natbib}
\usepackage{placeins}
\usepackage{authblk}

\usepackage{pgfplots}
\pgfplotsset{compat = 1.3}

\usepackage[accepted]{icml2016}

\icmltitlerunning{Texture Networks}

\newcommand{\fig}[1]{Figure~\ref{fig:#1}}
\newcommand{\sect}[1]{Sect.~\ref{sect:#1}}

\newcommand{\eq}[1]{(\ref{eq:#1})}

\newcommand{\eg}{\emph{e.g.}\xspace}

\newcommand{\xn}{{\ensuremath{\mathbf{x_0}}}\xspace}
\newcommand{\z}{{\ensuremath{\mathbf{z}}}\xspace}
\newcommand{\y}{{\ensuremath{\mathbf{y}}}\xspace}
\newcommand{\x}{{\ensuremath{\mathbf{x}}}\xspace}
\newcommand{\gen}{{\ensuremath{\mathbf{g}}}\xspace}
\newcommand{\loss}{{\ensuremath{\mathcal{L}}}\xspace}

\newcommand{\deflen}[2]{%
    \expandafter\newlength\csname #1\endcsname
    \expandafter\setlength\csname #1\endcsname{#2}%
}

\begin{document} 

\twocolumn[
\icmltitle{Texture Networks: Feed-forward Synthesis of Textures and Stylized Images}
\icmlauthor{Dmitry Ulyanov}{dmitry.ulyanov@skoltech.ru}
\icmladdress{Skolkovo Institute of Science and Technology \& Yandex, Russia}
\icmlauthor{Vadim Lebedev}{vadim.lebedev@skoltech.ru}
\icmladdress{Skolkovo Institute of Science and Technology \& Yandex, Russia}
\icmlauthor{Andrea Vedaldi}{vedaldi@robots.ox.ac.uk}
\icmladdress{University of Oxford, United Kingdom}
\icmlauthor{Victor Lempitsky}{lempitsky@skoltech.ru}
\icmladdress{ Skolkovo Institute of Science and Technology, Russia}
                   
\icmlkeywords{boring formatting information, machine learning, ICML}
\vskip 0.3in
]

\begin{abstract}
Gatys~et~al. recently demonstrated that deep networks can generate beautiful textures and stylized images from a single texture example. However, their methods requires a slow and memory-consuming optimization process. We propose here an alternative approach that moves the computational burden to a learning stage. Given a single example of a texture, our approach trains compact feed-forward convolutional networks to generate multiple samples of the same texture of arbitrary size and to transfer artistic style from a given image to any other image. The resulting networks are remarkably light-weight and can generate textures of quality comparable to Gatys~et~al., but hundreds of times faster. More generally, our approach highlights the power and flexibility of generative feed-forward models trained with complex and expressive loss functions.
\end{abstract} 

\section{Introduction}\label{submission}\label{s:intro}

Several recent works demonstrated the power of deep neural networks in the challenging problem of \textit{generating images}. Most of these proposed \textbf{generative networks} that produce images as output, using feed-forward calculations from a random seed; however, very impressive results were obtained by \cite{Gatys15a,Gatys15b} by using \textbf{networks descriptively}, as image statistics. Their idea is to reduce image generation to the problem of sampling at random from the set of images that match a certain statistics. In \textbf{texture synthesis}~\cite{Gatys15a}, the reference statistics is extracted from a single example of a visual texture, and the goal is to generate further examples of that texture. In \textbf{style transfer}~\cite{Gatys15b}, the goal is to match simultaneously the visual style of a first image, captured using some low-level statistics, and the visual content of a second image, captured using higher-level statistics. In this manner, the style of an image can be replaced with the one of another without altering the overall semantic content of the image.

\deflen{sixlen}{0.14\textwidth}

\begin{figure*}
    \includegraphics[width=\textwidth]{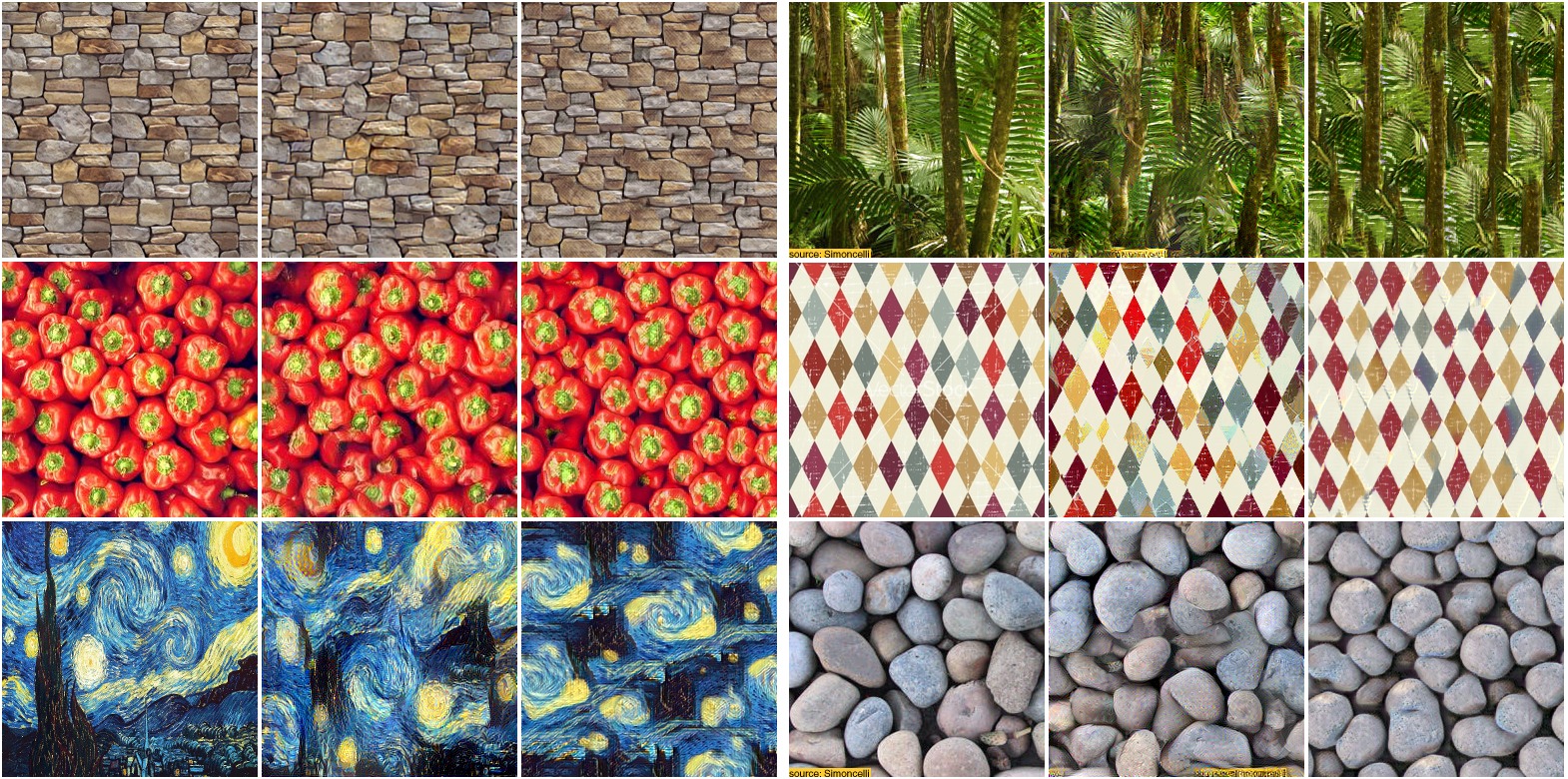}\\
    \begin{tabular}{cccccc}
    \makebox[\sixlen]{Input}&
    \makebox[\sixlen]{Gatys et al.}&
    \makebox[\sixlen]{Texture nets (ours)}\hspace{3mm}&
    \makebox[\sixlen]{Input}&
    \makebox[\sixlen]{Gatys et al.}&
    \makebox[\sixlen]{Texture nets (ours)}
    \end{tabular}
    \vspace{-0.5em}
    \caption{Texture networks proposed in this work are feed-forward architectures capable of learning to synthesize complex textures based on a single training example. The perceptual quality of the feed-forwardly generated textures is similar to the results of the closely related method suggested in  \cite{Gatys15a}, which use slow optimization process.}
    \label{fig:comparison_3way}
\end{figure*}

Matching statistics works well in practice, is conceptually simple, and demonstrates that off-the-shelf neural networks trained for generic tasks such as image classification can be re-used for image generation. However, the approach of \cite{Gatys15a,Gatys15b} has certain shortcomings too. Being based on an \emph{iterative optimization procedure}, it requires backpropagation to gradually change the values of the pixels until the desired statistics is matched. This iterative procedure requires several seconds in order to generate a relatively small image using a high-end GPU, while scaling to large images is problematic because of high memory requirements. By contrast, feed-forward generation networks can be expected to be much more efficient because they require a single evaluation of the network and do not incur in the cost of backpropagation.

In this paper we look at the problem of achieving the synthesis and stylization capability of descriptive networks using feed-forward generation networks. Our contribution is threefold. First, we show for the first time that a generative approach can produce textures of the quality and diversity comparable to the descriptive method. Second, we propose a generative method that is two orders of magnitude faster and one order of magnitude more memory efficient than the descriptive one. Using a single forward pass in networks that are remarkably compact make our approach suitable for video-related and possibly mobile applications. Third, we devise a new type of multi-scale generative architecture that is particularly suitable for the tasks we consider. 

The resulting fully-convolutional networks (that we call \textit{texture networks}) can  generate textures and process images of arbitrary size. Our approach also represents an interesting showcase of training conceptually-simple feed-forward architectures while using complex and expressive loss functions. We believe that other interesting results can be obtained using this principle.

The rest of the paper provides the overview of the most related approaches to image and texture generation (\sect{related}), describes our approach (\sect{method}), and provides extensive extensive qualitative comparisons on challenging textures and images (\sect{experiments}). 

\section{Background and related work}\label{sect:related}

\paragraph{Image generation using neural networks.} In general, one may look at the process of generating an image $\x$ as the problem of drawing a sample from a certain distribution $p(\x)$. In texture synthesis, the distribution is induced by an example texture instance $\xn$ (\eg~a polka dots image), such that we can write $\x \sim p(\x|\xn)$. In style transfer, the distribution is induced by an image $\xn$ representative of the visual style (\eg~an impressionist painting) and a second image $\x_1$ representative of the visual content (\eg~a boat), such that $\x \sim p(\x|\xn,\x_1)$.

\cite{Mahendran15,Gatys15a,Gatys15b} reduce this problem to the one of finding a \emph{pre-image} of a certain image statistics $\Phi(\x)\in\mathbb{R}^d$ and pose the latter as an optimization problem. In particular, in order to synthesize a texture from an example image $\xn$, the pre-image problem is:
\begin{equation}\label{eq:synth}
   \operatornamewithlimits{argmin}_{\x\in\mathcal{X}} \| \Phi(\x) - \Phi(\xn) \|^2_2.
\end{equation}
Importantly, the pre-image $\x: \Phi(\x) \approx \Phi(\xn)$ is usually not unique, and sampling pre-images achieves diversity. In practice, samples are extracted using a local optimization algorithm $\mathcal{A}$ starting from a random initialization $\z$. Therefore, the generated image is the output of the function
\begin{equation}\label{e:snyth-real}
\operatornamewithlimits{localopt}_{\x\in\mathcal{X}}(\| \Phi(\x) - \Phi(\xn) \|^2_2; \mathcal{A},\z),
\quad
\z \sim \mathcal{N}(\mathbf{0},\Sigma).
\end{equation}
This results in a distribution $p(\x|\xn)$ which is difficult to characterise, but is easy to sample and, for good statistics $\Phi$, produces visually pleasing and diverse images. Both \cite{Mahendran15} and \cite{Gatys15a,Gatys15b} base their statistics on the response that $\x$ induces in deep neural network layers. Our approach reuses in particular the statistics based on correlations of convolutional maps proposed by \cite{Gatys15a,Gatys15b}.

\paragraph{Descriptive texture modelling.} The approach described above has strong links to many well-known models of visual textures. For texture, it is common to assume that $p(\x)$ is a stationary \emph{Markov random field} (MRF). In this case, the texture is ergodic and one may considers local spatially-invariant statistics $\psi \circ F(\x;i), i\in\Omega$, where $i$ denotes a spatial coordinate. Often $F$ is the output of a bank of linear filters and $\psi$ an histogramming operator. Then the spatial average of this local statistics on the prototype texture $\xn$ approximates its sample average
\begin{equation}\label{eq:frame}
\phi(\xn) 
= \frac{1}{|\Omega|} \sum_{i=1}^{|\Omega|}
\psi \circ F(\xn;i)
\approx \underset{\x \sim p(\x)}{E}[\psi \circ F_l(\x;0)].
\end{equation}
The FRAME model of~\cite{Zhu98} uses this fact to induce the maximum-entropy distribution over textures
$
  p(\x) \propto \exp(- \langle \lambda, \phi(\x)\rangle),
$
where $\lambda$ is a parameter chosen so that the marginals match their empirical estimate, i.e.
$
 E_{\x \sim p(\x)}[\phi(\x)] = \phi(\xn).
$

A shortcoming of FRAME is the difficulty of sampling from the maxent distribution. \cite{Portilla00} addresses this limitation by proposing to directly find images $\x$ that match the desired statistics $\Phi(\x) \approx \Phi(\xn)$, pioneering the pre-image method of~\eq{synth}.

Where~\cite{Zhu98,Portilla00} use linear filters, wavelets, and histograms to build their texture statistics, \cite{Mahendran15,Gatys15a,Gatys15a} extract statistics from pre-trained deep neural networks. \cite{Gatys15b} differs also in that it considers the style transfer problem instead of the texture synthesis one.

\paragraph{Generator deep networks.} An alternative to using a neural networks as descriptors is to construct generator networks $\x = \gen(\z)$ that produce \emph{directly} an image $\x$ starting from a vector of random or deterministic parameters $\z$.

Approaches such as \cite{Dosovitskiy15} learn a mapping from deterministic parameters $\z$ (\eg the type of object imaged and the viewpoint) to an image $\x$. This is done by fitting a neural network to minimize the discrepancy $\|\x_i - \gen(\z_i)\|$ for known image-parameter pairs $(\x_i,\z_i)$. While this may produce visually appealing results, it requires to know the relation $(\x,\z)$ beforehand and cannot express any diversity beyond the one captured by the parameters.

An alternative is to consider a function $\gen(\z)$ where the parameters $\z$ are unknown and are sampled from a (simple) random distribution. The goal of the network is to map these random values to plausible images $\x = \gen(\z)$. This requires measuring the quality of the sample, which is usually expressed as a distance between $\x$ and a set of example images $\x_1,\dots,\x_n$. The key challenge is that the distance must be able to generalize significantly from the available examples in order to avoid penalizing sample diversity.

Generative Adversarial Networks (GAN;~\cite{Goodfellow14}) address this problem by training, together with the generator network $\gen(\z)$, a second \emph{adversarial} network $f(\x)$ that attempts to distinguish between samples $\gen(\z)$ and natural image samples. Then $f$ can be used as a measure of quality of the samples and \gen can be trained to optimize it. LAPGAN~\cite{Denton15} applies GAN to a Laplacian pyramid of convolutional networks and DCGAN~\cite{Radford15} further optimizes GAN and learn is from very large datasets.

\paragraph{Moment matching networks.} The maximum entropy model of~\cite{Zhu98} is closely related to the idea of \emph{Maximum Mean Discrepancy} (MMD) introduced in \cite{Gretton06}. Their key observation  the expected value $\mu_p = E_{\x \sim p(\x)}[\phi(\x)]$ of certain statistics $\phi(\x)$ uniquely identifies the distribution $p$. \cite{Li15,Dziugaite15} derive from it a loss function alternative to GAN by comparing the statistics averaged over network samples $\frac{1}{m}\sum_{i=1}^m \phi \circ \gen(\z_i)$ to the statistics averaged over empirical samples $\frac{1}{m}\sum_{i=1}^m \phi(\x_i)$. They use it to train a \emph{Moment Matching Network} (MMN) and apply it to generate small images such as MNIST digits. Our networks are similar to moment matching networks, but use very specific statistics and applications quite different from the considered in \cite{Li15,Dziugaite15}.
\section{Texture networks}\label{sect:method}

\begin{figure*}
    \centering
    \includegraphics[width=\textwidth]{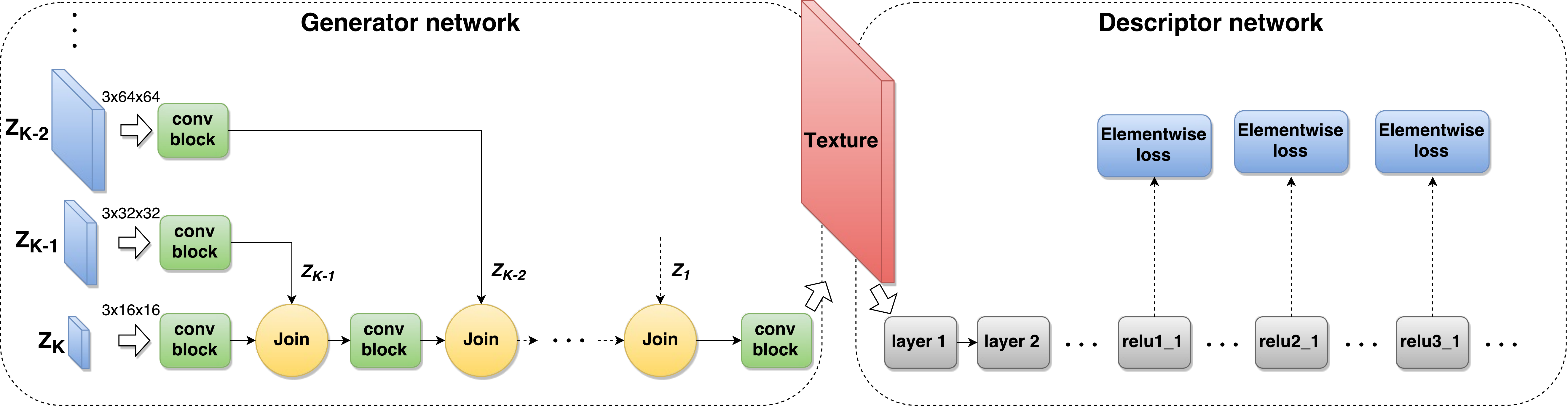}
    \caption{Overview of the proposed architecture (texture networks). We train a \textit{generator network} (left) using a powerful loss based on the correlation statistics inside a fixed pre-trained \emph{descriptor network} (right). Of the two networks, only the generator is updated and later used for texture or image synthesis. The \texttt{conv} block contains multiple convolutional layers and non-linear activations and the \texttt{join} block upsampling and channel-wise concatenation. Different branches of the generator network operate at different resolutions and are excited by noise tensors $z_i$ of different sizes.}
    \label{fig:arch}
\end{figure*}


We now describe the proposed method in detail. At a high-level (see~\fig{arch}), our approach is to train a feed-forward \textbf{generator network} $\gen$ which takes a noise sample $\z$ as input and produces a texture sample $\gen(\z)$ as output. For style transfer, we extend this \emph{texture network} to take both a noise sample $\z$ and a content image $\y$ and then output a new image $\gen(\y,\z)$ where the texture has been applied to $\y$ as a visual style. A separate generator network is trained for each texture or style and, once trained, it can synthesize an arbitrary number of images of arbitrary size in an efficient, feed-forward manner. 

A key challenge in training the generator network $\gen$ is to construct a loss function that can assess automatically the quality of the generated images. For example, the key idea of GAN is to \emph{learn} such a loss along with the generator network. We show in Sect.~\ref{sect:loss} that a very powerful loss can be derived from pre-trained and fixed \textbf{descriptor networks} using the statistics introduced in \cite{Gatys15a,Gatys15b}. Given the loss, we then discuss the architecture of the generator network for texture synthesis (Sect.~\ref{sect:gen}) and then generalize it to style transfer (Sect~\ref{sect:transfer}).

\subsection{Texture and content loss functions}\label{sect:loss}

Our loss function is derived from~\cite{Gatys15a,Gatys15b} and compares image statistics extracted from a fixed pre-trained descriptor CNN (usually one of the VGG CNN~\cite{Simonyan14,Chatfield14} which are pre-trained for image  classification on the ImageNet ILSVRC 2012 data). The descriptor CNN is used to measure the mismatch between the \textit{prototype} texture \xn and the generated image \x. Denote by $F^l_i(\x)$ the $i$-th map (feature channel) computed by the $l$-th convolutional layer by the descriptor CNN applied to image \x. The \emph{Gram matrix} $G^l(\x)$ is defined as the matrix of scalar (inner) products between such feature maps:
\begin{equation}\label{eq:gram}
    G^l_{ij}(\x) = \langle F^l_i(\x),F^l_j(\x)\rangle\,.
\end{equation}
Given that the network is convolutional, each inner product implicitly sums the products of the activations of feature $i$ and $j$ at all spatial locations, computing their (unnormalized) empirical correlation. Hence $G^l_{ij}(\x)$ has the same general form as \eq{frame} and, being an orderless statistics of local stationary features, can be used as a texture descriptor.

In practice, \cite{Gatys15a,Gatys15b} use as texture descriptor the combination of several Gram matrices $G^l, l\in L_T$, where $L_T$ contains selected indices of convolutional layer in the descriptor CNN. This induces the following \textit{texture loss} between images \x and \xn:
\begin{equation}\label{eq:textloss}
    \loss_T(\x;\xn) = \sum_{l \in L_T} \|G^l(\x) - G^l(\xn) \|_2^2\,.
\end{equation}

In addition to the texture loss \eq{textloss}, \cite{Gatys15b} propose to use as \emph{content loss} the one introduced by \cite{Mahendran15}, which compares images based on the output $F^l_i(\x)$ of certain convolutional layers $l\in L_C$ (without computing further statistics such as the Gram matrices). In formulas
\begin{equation}\label{eq:contentloss}
    \loss_C(\x;\y) = \sum_{l \in L_C} \sum_{i=1}^{N_l} \|F^l_i(\x) - F^l_i(\y) \|_2^2\,,
\end{equation}
where $N_l$ is the number of maps (feature channels) in layer $l$ of the descriptor CNN. The key difference with the texture loss~\eq{textloss} is that the content loss compares feature activations at corresponding spatial locations, and therefore preserves spatial information. Thus this loss is suitable for content information, but not for texture information.

Analogously to \cite{Gatys15a}, we use the texture loss \eq{textloss} alone when training a generator network for texture synthesis, and we use a weighted combination of the texture loss \eq{textloss} and the content loss \eq{contentloss} when training a generator network for stylization.
In the latter case, the set $L_C$ does not includes layers as shallow as the set $L_T$ as only the high-level content should be preserved.

\begin{figure*}
\begin{center}
\includegraphics[width=0.8\textwidth]{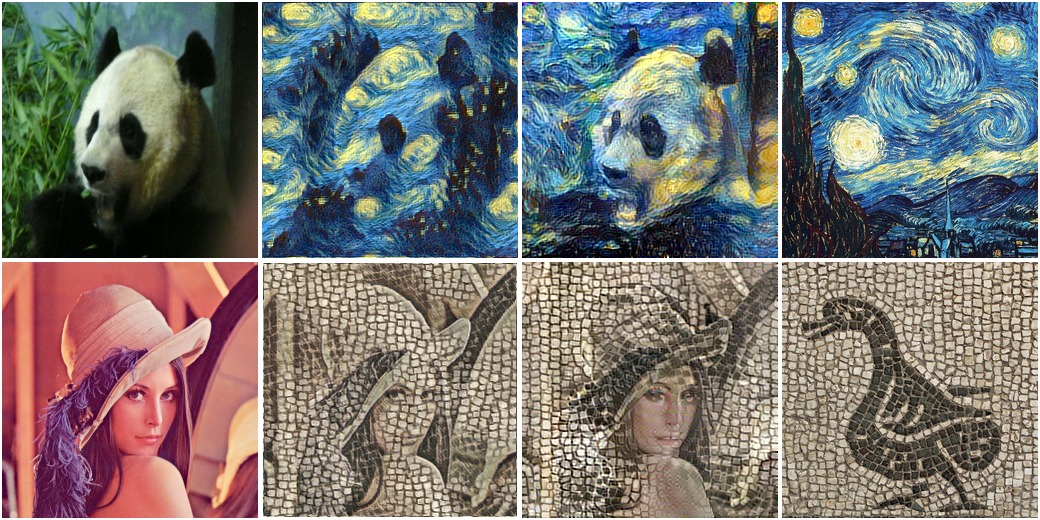}\\
\makebox[0.2\textwidth]{Content}
\makebox[0.2\textwidth]{Texture nets (ours)}
\makebox[0.2\textwidth]{Gatys et al.}
\makebox[0.2\textwidth]{Style}\\
\end{center}
\vspace{-1em}
\caption{Our approach can also train feed-forward networks to transfer style from artistic images (left). After training, a network can transfer the style to any new image (e.g.\ right) while preserving semantic content. For some styles (bottom row), the perceptual quality of the result of our feed-forward transfer is comparable with the optimization-based method \cite{Gatys15b}, though for others the results are not as impressive (top row and \cite{supmat}).}
\label{fig:optim_comparison}
\end{figure*}

\subsection{Generator network for texture synthesis}\label{sect:gen}

We now discuss the architecture and the training procedure for the generator network $\gen$ for the task of texture synthesis. We denote the parameters of the generator network as $\theta$. The network is trained to transform a noise vector $\z$ sampled from a certain distribution $\cal Z$ (which we set to be uniform i.i.d.) into texture samples that match, according to the texture loss~\eq{textloss}, a certain prototype texture \xn:
\begin{equation}\label{eq:textlearn}
\theta_\xn = \operatornamewithlimits{argmin}_\theta
\;
E_{\z\sim \cal Z}\,\left[\, \loss_T\left(\gen(\z;\theta),\,\xn\right) \,\right]\,.
\end{equation}

\paragraph{Network architecture.} We experimented with several architectures for the generator network $\gen$. The simplest are chains of convolutional, non-linear activation, and upsampling layers that start from a noise sample \z in the form of a small feature map and terminate by producing an image. While models of this type produce reasonable results, we found that \emph{multi-scale architectures} result in images with smaller texture loss and better perceptual quality while using fewer parameters and training faster. Figure~\ref{fig:arch} contains a high-level representation of our reference multi-scale architecture, which we describe next.

The reference texture $\xn$ is a tensor $\mathbb{R}^{M\times M\times 3}$ containing three color channels. For simplicity, assume that the spatial resolution $M$ is a power of two. The input noise $\z$ comprises $K$ random tensors $\z_i \in \mathbb{R}^{\frac{M}{2^i} \times \frac{M}{2^i}}$, $i = 1, 2, \dots, K$ (we use $M = 256$ and $K=5$) whose entries are i.i.d.\ sampled from a uniform distribution. Each random noise tensor is first processed by a sequence of convolutional and non-linear activation layers, then upsampled by a factor of two, and finally concatenated as additional feature channels to the partially processed tensor from the scale below. The last full-resolution tensor is ultimately mapped to an RGB image \x by a bank of $1\times 1$ filters.

Each convolution block in Figure~\ref{fig:arch} contains three convolutional layers, each of which is followed by a ReLU activation layer. The convolutional layers contain respectively $3\times 3$, $3\times 3$ and $1\times 1$ filters. Filers are computed densely (stride one) and applied using circular convolution to remove boundary effects, which is appropriate for textures. The number of feature channels, which equals the number of filters in the preceding bank, grows from a minimum of $8$ to a maximum of $40$. The supplementary material specifies in detail the network configuration which has only $\sim$65K parameters, and can be compressed to $\sim$300 Kb of memory.

Upsampling layers use simple nearest-neighbour interpolation (we also experimented strided full-convolution~\cite{Long15,Radford15}, but the results were not satisfying). We found that training benefited significantly from inserting batch normalization layers~\cite{Ioffe15} right after each convolutional layer and, most importantly, right before the concatenation layers, since this balances gradients travelling along different branches of the network. 

\paragraph{Learning.} Learning optimizes the objective \eq{textlearn} using stochastic gradient descent (SGD). At each iteration, SGD draws a mini-batch of noise vectors $\z_k,k=1,\dots,B$, performs forward evaluation of the generator network to obtained the corresponding images  $\x_k=\gen(\z_k,\theta)$, performs forward evaluation of the descriptor network to obtain Gram matrices $G^l(\x_k),l\in L_T$, and finally computes the loss \eq{textloss} (note that the corresponding terms $G^l(\xn)$ for the reference texture are constant). After that, the gradient of the texture loss with respect to the generator network parameters $\theta$ is computed using backpropagation, and the gradient is used to update the parameters. Note that LAPGAN~\cite{Denton15} also performs multi-scale processing, but uses layer-wise training, whereas our generator is trained end-to-end.

\begin{figure*}
\deflen{fivelen}{0.175\textwidth}
\centering
\includegraphics[width=\textwidth]{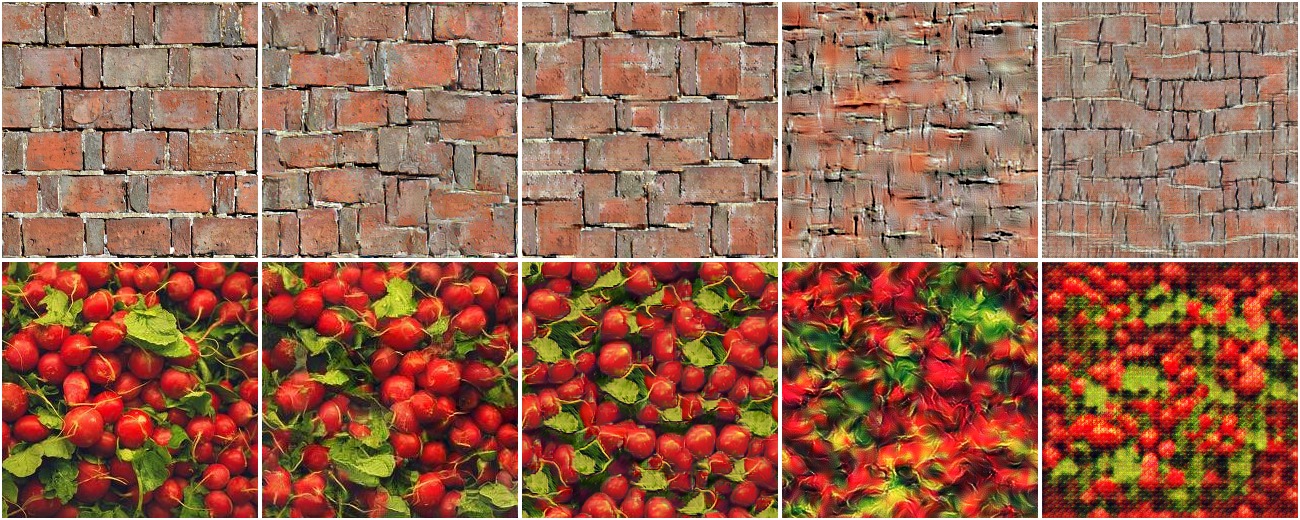}
\begin{tabular}{ccccc}
    \makebox[\fivelen]{Input}&
    \makebox[\fivelen]{Gatys et al.}&
    \makebox[\fivelen]{Texture nets (ours)}&
    \makebox[\fivelen]{Portilla, Simoncelli}&
    \makebox[\fivelen]{DCGAN}
    \end{tabular}
    \\
\caption{Further comparison of textures generated with several methods including the original statistics matching method  \cite{Portilla00} and the DCGAN \cite{Radford15} approach. Overall, our method and \cite{Gatys15a} provide better results, our method being hundreds times faster.}
\label{fig:comparison}
\end{figure*}

\subsection{Style transfer}\label{sect:transfer}

In order to extend the method to the task of image stylization, we make several changes. Firstly, the generator network $\x=\gen(\y,\z;\theta)$ is modified to take as input, in addition to the noise variable $\z$, the image $\y$ to which the noise should be applied. The generator network is then trained to output an image \x that is close in content to \y and in texture/style to a reference texture \xn. For example, \y could be a photo of a person, and \xn an impressionist painting.

\paragraph{Network architecture.} The architecture is the same as the one used for texture synthesis with the important difference that now the noise tensors $\z_i,i=1,\dots,K$ at the $K$ scales are concatenated (as additional feature channels) with downsampled versions of the input image \y. For this application, we found beneficial to increased the number of scales from $K=5$ to $K=6$. 

\paragraph{Learning.} Learning proceeds by sampling noise vectors $\z_i\sim\mathcal{Z}$ and natural images $\y_i\sim\mathcal{Y}$ and then adjusting the parameters $\theta$ of the generator $\gen(\y_i,\z_i;\theta)$ in order to minimize the combination of content and texture loss:
\begin{align} \label{eq:imlearn}
\theta_\xn = \operatornamewithlimits{argmin}_\theta \; E_{\z\sim \cal Z;\; \y \sim \cal Y}\,[&\\ \loss_T\left(\gen(\y,\z;\theta),\,\xn\right) +\,& \nonumber \alpha\,\loss_C \left(\gen(\y,\z;\theta),\,\y\right) \,]\,.
\end{align}
Here $\mathcal{Z}$ is the same noise distribution as for texture synthesis, $\mathcal{Y}$ empirical distribution on naturals image (obtained from any image collection), and  $\alpha$ a parameter that trades off preserving texture/style and content. In practice, we found that learning is surprisingly resilient to overfitting and that it suffices to approximate the distribution on natural images  $\mathcal{Y}$ with a very small pool of images (e.g.\ $16$). In fact, our qualitative results degraded using too many example images. We impute this to the fact that stylization by a convolutional architecture uses local operations; since the same local structures exist in different combinations and proportions in different natural images \y, it is difficult for local operators to match in all cases the overall statistics of the reference texture \xn, where structures exist in a fixed arbitrary proportion. Despite this limitation, the perceptual quality of the generated stylized images is usually very good, although for some styles we could not match the quality of the original stylization by optimization of~\cite{Gatys15b}.

\section{Experiments}
\label{sect:experiments}

\begin{figure*}
\centering
\includegraphics[width=0.8\textwidth]{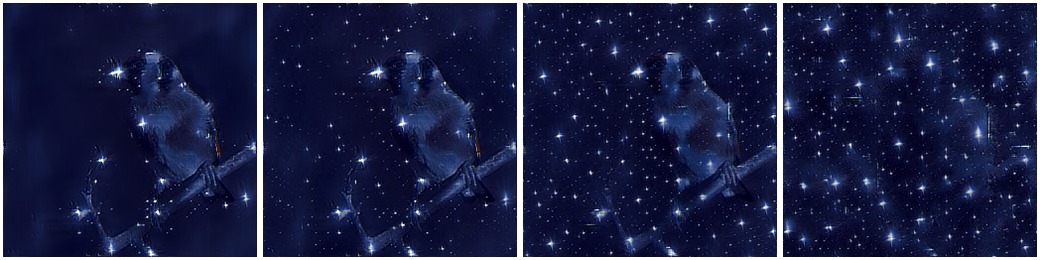}\\
\makebox[0.2\textwidth]{$k=0.01$}%
\makebox[0.2\textwidth]{$k=0.1$}%
\makebox[0.2\textwidth]{$k=1$}%
\makebox[0.2\textwidth]{$k=10$}\\
\caption{Our architecture for image stylization takes the content image and the noise vector as inputs. By scaling the input noise by different factors $k$ we can affect the balance of style and content in the output image without retraining the network.}
\label{fig:noise_levels}
\vspace{-3mm}
\end{figure*}

\paragraph{Further technical details.} The generator network weights were initialized using Xavier's method. Training used \emph{Torch7}'s implementation of Adam~\cite{Kingma14}, running it for $2000$ iteration. The initial learning rate of $0.1$ was reduced by a factor $0.7$ at iteration $1000$ and then again every $200$ iterations. The batch size was set to $16$. Similar to \cite{Gatys15a}, the texture loss uses the layers $L_T=\{\texttt{relu1\_1}, \texttt{relu2\_1}, \texttt{relu3\_1}, \texttt{relu4\_1}, \texttt{relu5\_1}\}$ of VGG-19 and the content loss the layer $L_C = \{\texttt{relu4\_2}\}$. Fully training a single model required just two hours on an NVIDIA Tesla K40, and visually appealing results could be obtained much faster, after just a few epochs.





\paragraph{Texture synthesis.} We compare our method to \cite{Gatys15a,Gatys15b} using the popular implementation of~\cite{Johnson14}, which produces comparable if not better results than the implementation eventually released by the authors. We also compare to the DCGAN~\cite{Radford15} version of adversarial networks \cite{Goodfellow14}. Since DCGAN training requires multiple example images for training, we extract those as sliding $64\times 64$ patches from the $256 \times 256$ reference texture \xn; then, since DCGAN is fully convolutional, we use it to generate larger $256 \times 256$ images simply by inputting a larger noise tensor. Finally, we compare to \cite{Portilla00}.

\fig{comparison} shows the results obtained by the four methods on two challenging textures of \cite{Portilla00}. Qualitatively, our generator CNN and \cite{Gatys15a}'s results are comparable and superior to the other methods; however, the generator CNN is much more efficient (see Sect.~\ref{sect:speed}). \fig{comparison_3way} includes further comparisons between the generator network and \cite{Gatys15a} and many others are included in the supplementary material.

\paragraph{Style transfer.} For training, example natural images were extracted at random from the ImageNet ILSVRC 2012 data. As for the original method of~\cite{Gatys15b}, we found that style transfer is sensitive to the trade-off parameter $\alpha$ between texture and  content loss in \eq{contentloss}. At test time this parameter is not available in our method, but we found that the trade-off can still be adjusted by changing the magnitude of the input noise \z (see~\fig{noise_levels}).

\begin{figure*}
\centering
\includegraphics[width=\textwidth]{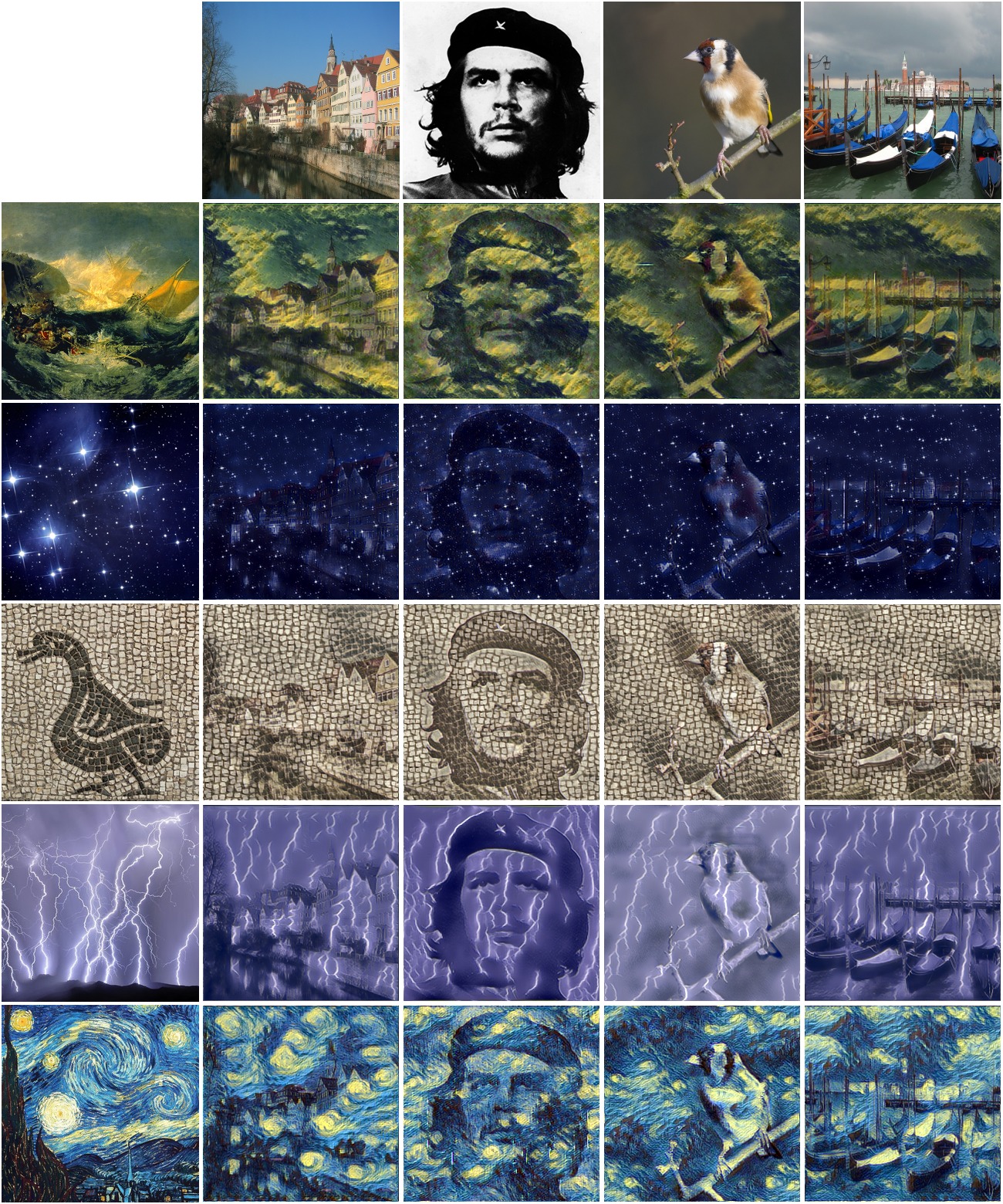}
\caption{Stylization results for various styles and inputs (one network per row). Our approach can handle a variety of styles. The generated images are of \texttt{256x256} resolution and are computed in about 20 milliseconds each.}
\label{fig:style_examples}
\end{figure*}

We compared our method to the one of \cite{Gatys15b,Johnson14} using numerous style and content images, including the ones in \cite{Gatys15b}, and found that results are qualitatively comparable. Representative comparisons (using a fixed parameter $\alpha$) are included in \fig{optim_comparison} and many more in the supplementary material. Other qualitative results are reported in \fig{style_examples}.

\subsection{Speed and memory}\label{sect:speed}

\begin{figure}
    \centering
    \newlength\figureheight
    \newlength\figurewidth
    \setlength\figureheight{0.35\textwidth}
    \setlength\figurewidth{0.45\textwidth}
    \input{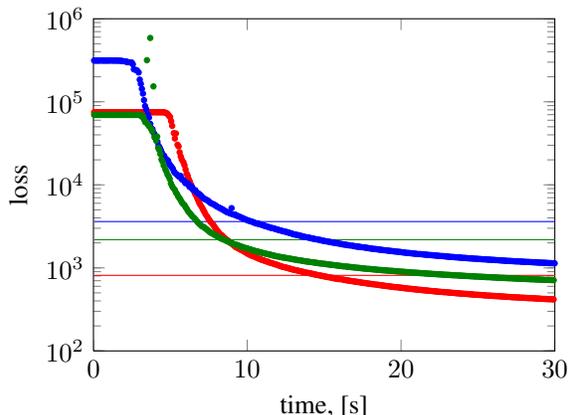}
    \vspace{-5mm}
    \caption{The objective values (log-scale) within the optimization-based method \cite{Gatys15a} for three randomly chosen textures are plotted as functions of time. Horizontal lines show the style loss achieved by our feedforward algorithm (mean over several samples) for the same textures. It takes the optimization within \cite{Gatys15a} around $10$ seconds ($500$x slower than feedforward generation) to produce samples with comparable loss/objective. }
    \label{fig:time}
\end{figure}

We compare quantitatively the speed of our method and of the iterative optimization of \cite{Gatys15a} by measuring how much time it takes for the latter and for our generator network to reach a given value of the loss $\loss_T(\x,\xn)$. \fig{time} shows that iterative optimization requires about $10$ seconds to generate a sample $\x$ that has a loss comparable to the output $\x = \gen(\z)$ of our generator network. Since an evaluation of the latter requires $\sim$20ms, we achieve a $500\times$ speed-up, which is sufficient for \emph{real-time applications} such as video processing. There are two reasons for this significant difference: the generator network is much smaller than the VGG-19 model evaluated at each iteration of~\cite{Gatys15a}, and our method requires a single network evaluation. By avoiding backpropagation, our method also uses significantly less memory (170 MB to generate a $256\times 256$ sample, vs 1100 MB of \cite{Gatys15a}).

\section{Discussion}

We have presented a new deep learning approach for texture synthesis and image stylization. Remarkably, the approach is able to generate complex textures and images in a purely feed-forward way, while matching the texture synthesis capability of \cite{Gatys15a}, which is based on multiple forward-backward iterations. In the same vein as \cite{Goodfellow14,Dziugaite15,Li15}, the success of this approach  highlights the suitability of feed-forward networks for complex data generation and for solving complex tasks in general. The key to this success is the use of complex loss functions that involve different feed-forward architectures serving as ``experts'' assessing the performance of the feed-forward generator. 

While our method generally obtains very good result for texture synthesis, going forward we plan to investigate better stylization losses to achieve a stylization quality comparable to~\cite{Gatys15b} even for those cases (\eg~\fig{optim_comparison}.top) where our current method achieves less impressive results.


\FloatBarrier
\bibliography{texture_nets_main}
\bibliographystyle{icml2016}

\pagebreak
\section*{Supplementary material} 

Below, we provide several additional qualitative results demonstrating the performance of texture networks and comparing them to Gatys~et~al.~approaches.

\subsection{A note on generator architecture}
\begin{figure}[H]
    \includegraphics[width=0.9\linewidth]{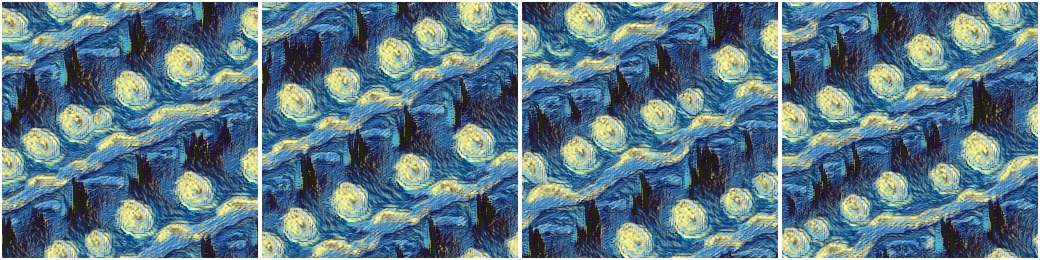}
    \caption{The example of an overfitted generator. Although every "star" differs from each other in details you may notice a common structure. This also illustrates generators love (when overfitted) for diagonal, horizontal and vertical lines as \texttt{3x3} filters are used.}
    \label{fig:degrading}
\end{figure}

Since the generator is only restricted to produce good images in terms of texture loss, nothing stops it from generating samples with small variance between them. Therefore, if a model archives lower texture loss it does not implicate that this model is preferable. The generator should be powerful enough to combine texture elements, but not too complex to degrade to similar samples. If degrading effect is noticed the noise amount increasing can help in certain cases. \fig{degrading} shows a bad case, with too much iteration performed. This degrading effect is similar to overfitting but there is no obvious way to control it as there is no analogue of validation set available. 

\begin{center}
    \includegraphics[width=\textwidth]{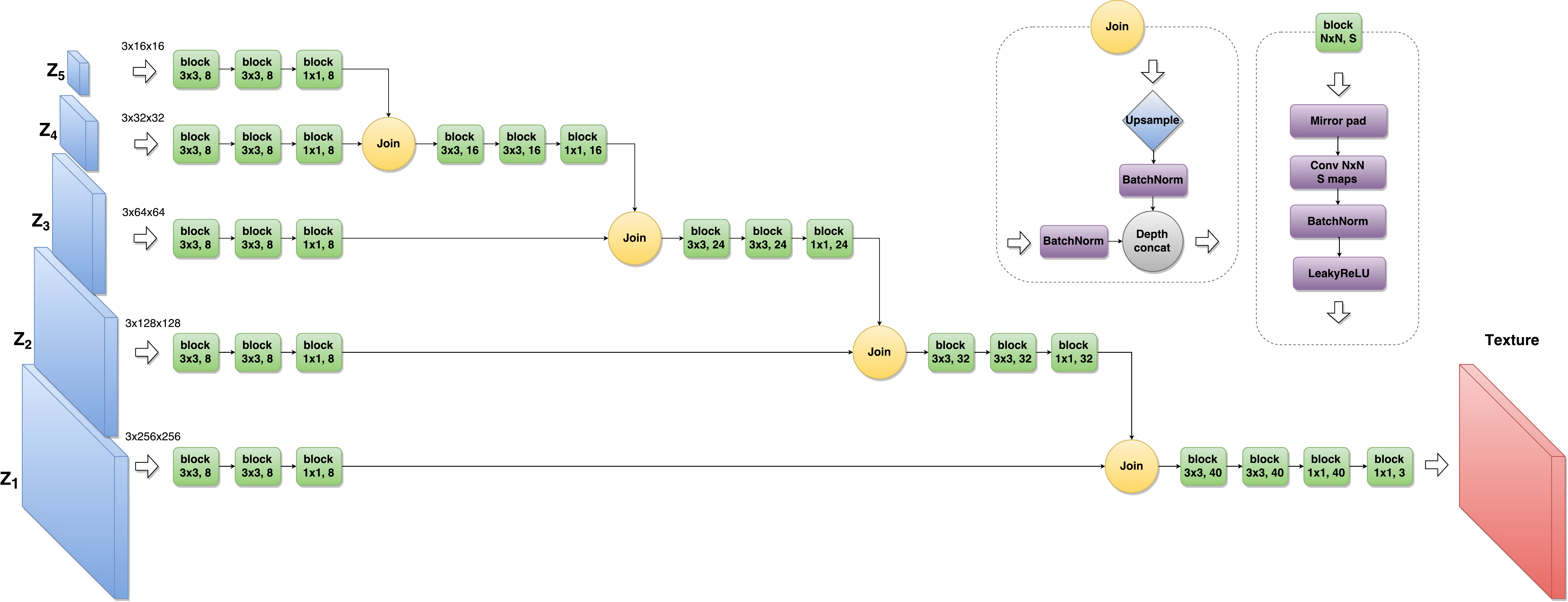}
    \captionof{figure}{A more exact scheme of the architecture used for texture generation in our experiments.}
\end{center}

\newpage

\begin{figure*}
    \includegraphics[width=\textwidth]{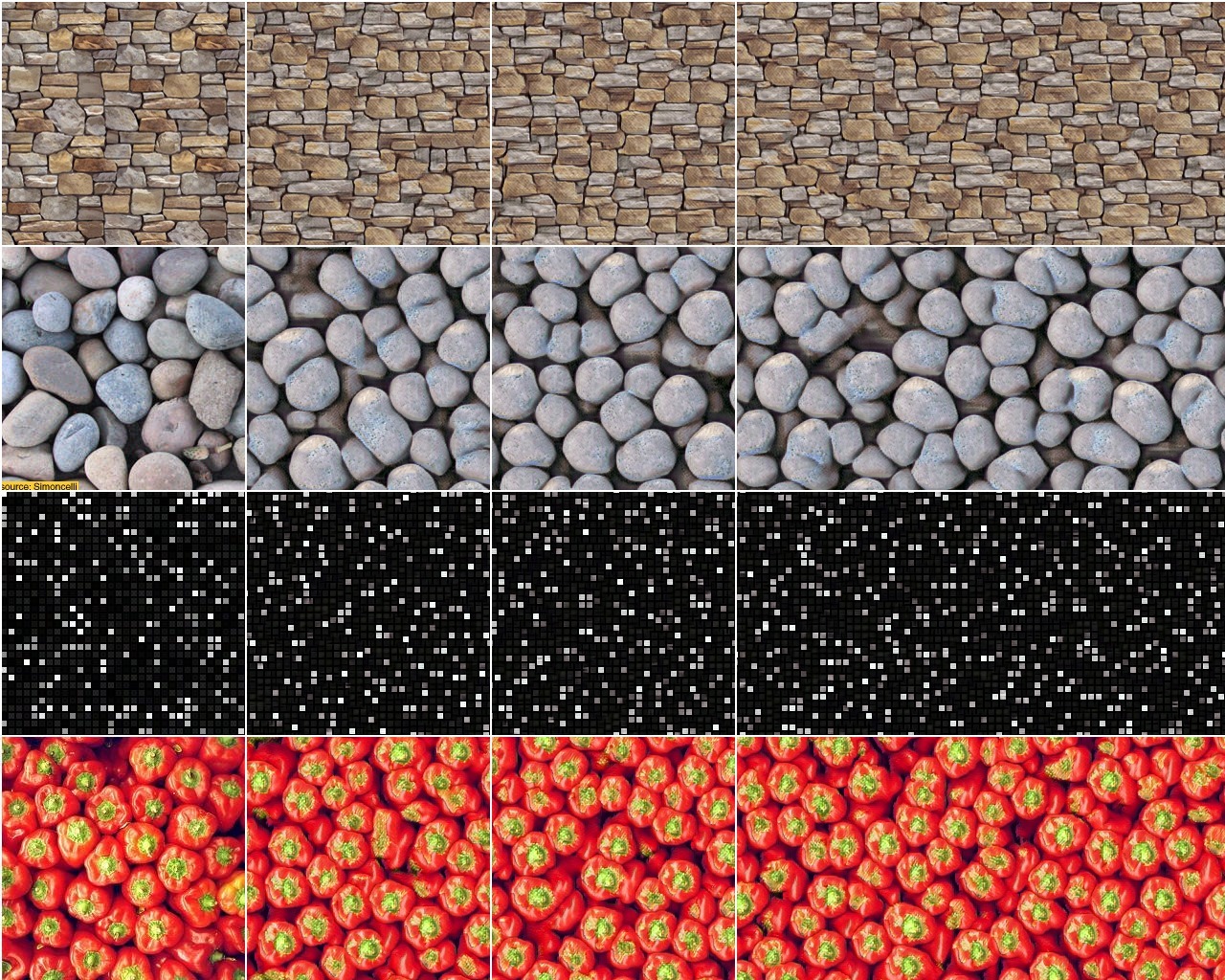}\\
    \begin{tabular}{cccc}
    \makebox[\fivelen]{Input}&
    \makebox[\fivelen]{Sample 1}&
    \makebox[\fivelen]{Sample 2}&\hspace{15mm}
    \makebox[\fivelen]{Sample 512x256}
    \end{tabular}\\
    \caption{Various samples drawn from three texture networks trained for the samples shown on the left. While training was done for \texttt{256x256} samples, in the right column we show the generated textures for a different resolution.}
    \label{fig:samples1}
\end{figure*}

\begin{figure*}
    \includegraphics[width=\textwidth]{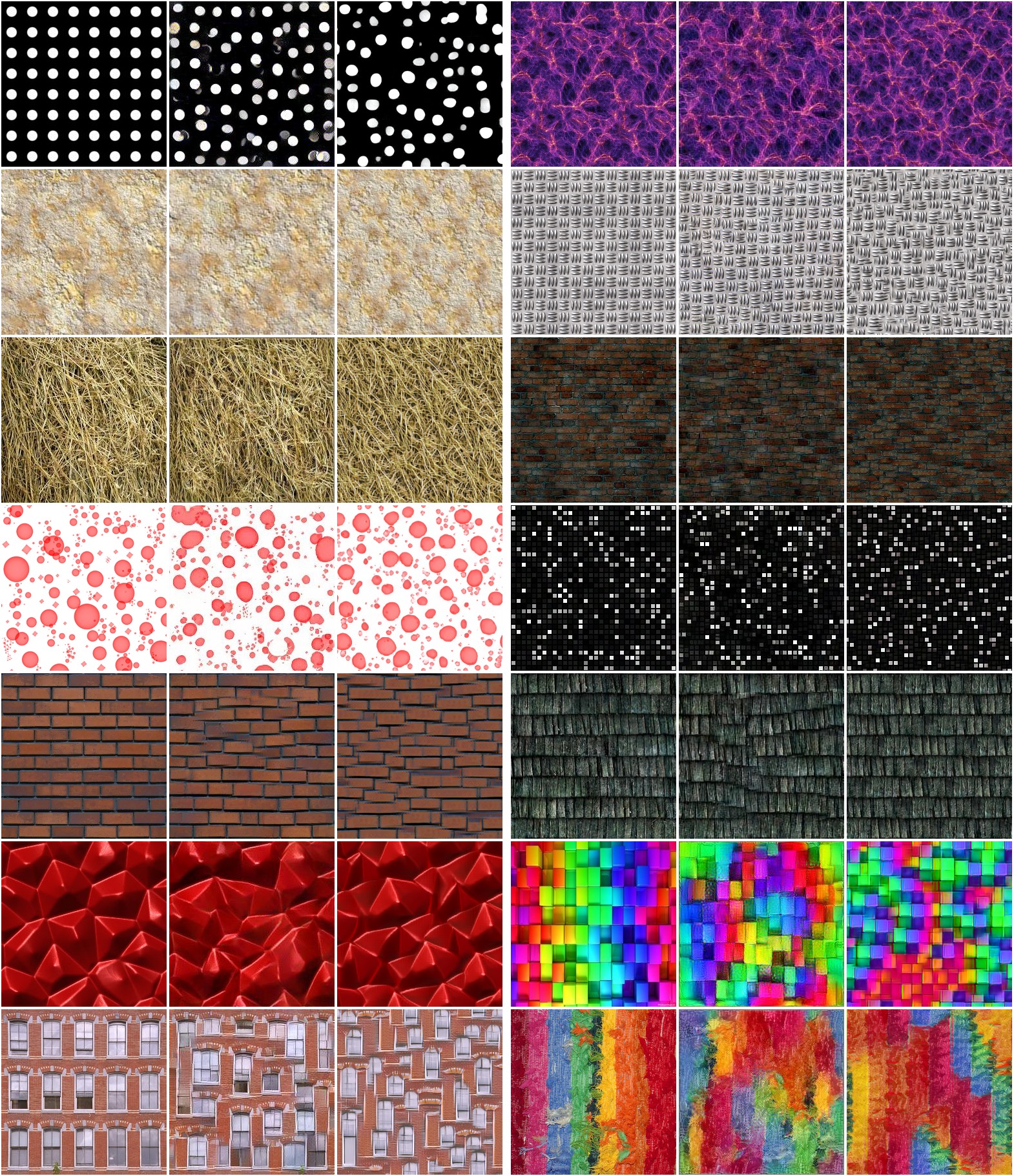}\\
    \begin{tabular}{cccccc}
    \makebox[\sixlen]{Input}&
    \makebox[\sixlen]{Gatys et al.}&
    \makebox[\sixlen]{Texture nets}\hspace{3mm}&
    \makebox[\sixlen]{Input}&
    \makebox[\sixlen]{Gatys et al.}&
    \makebox[\sixlen]{Texture nets}
    \end{tabular}
    \\
    \caption{More comparison with Gatys~et~al.\ for texture synthesis.}
    \label{fig:comparison_sup}
\end{figure*}

\begin{figure*}
    \includegraphics[width=\textwidth]{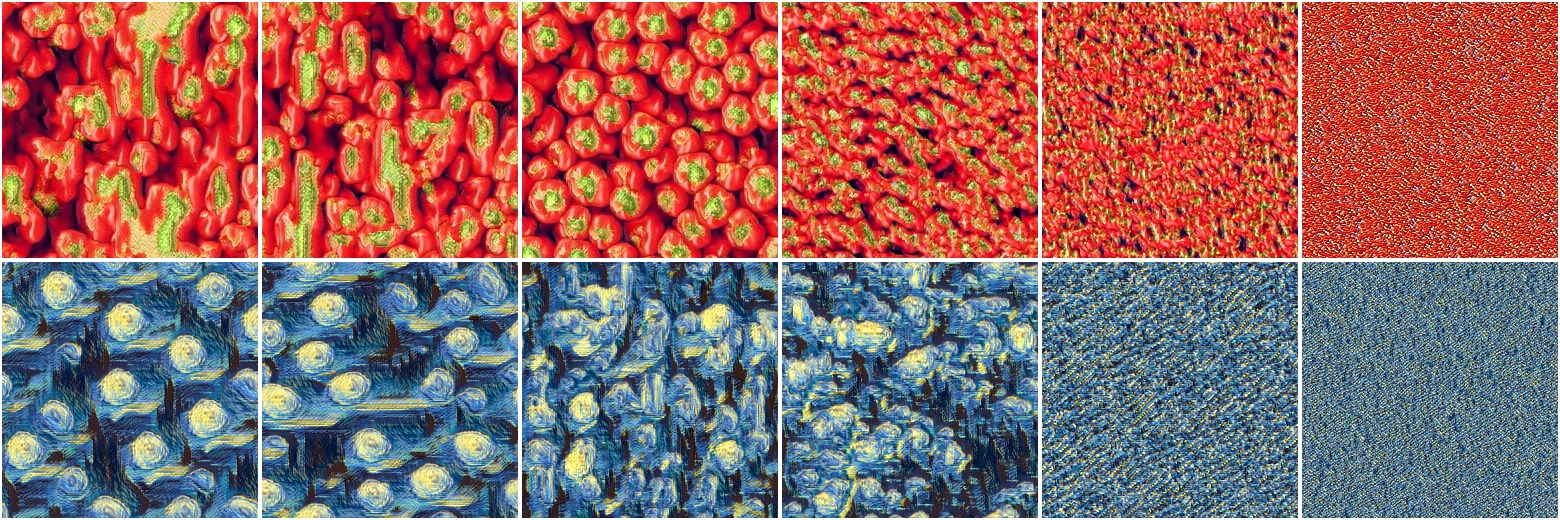}
    \caption{Ablation example: here we generate examples from the pretrained networks after zeroing out all inputs to the multi-scale architecture except for one scale. This let us analyze the processes inside the generator. For peppers image we observe that depth $K=4$ is enough for generator to perform well. Texture elements in starry night are bigger therefore the deepest input blob is used. Note that the generator is limited by the VGG network capacity and cannot capture larger texture elements than the receptive field of the last convolution layer.}
    \label{fig:zero_all_but_one}
\end{figure*}


\begin{figure*}
\centering
\includegraphics[width=\textwidth]{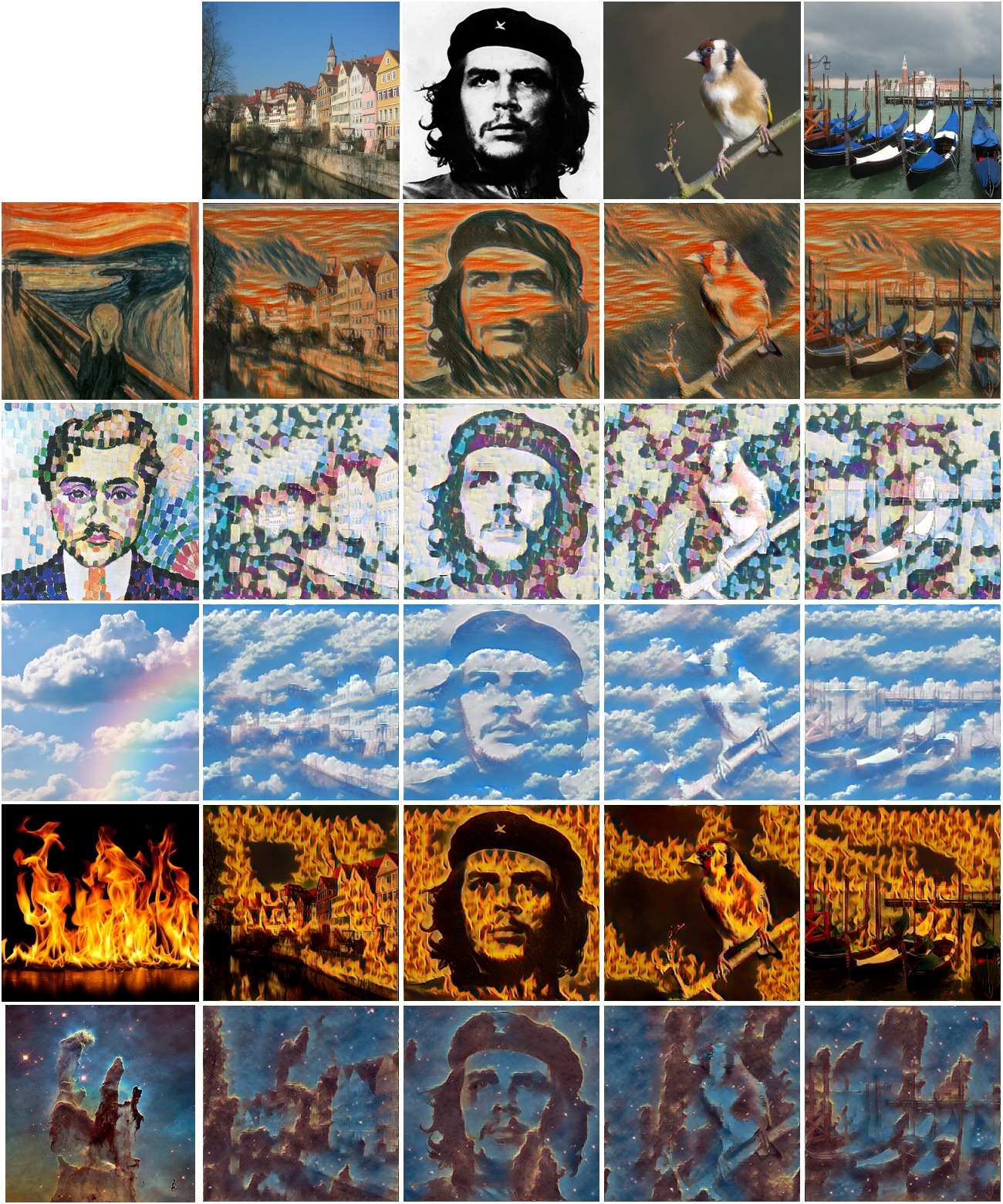}
\caption{Style transfer results for more styles. As before, each row represents a network trained with particular style with the original style image on the left, and columns correspond to different content images, with the original image on the top.}
\end{figure*}

\begin{figure*}
\centering
\includegraphics[width=\textwidth]{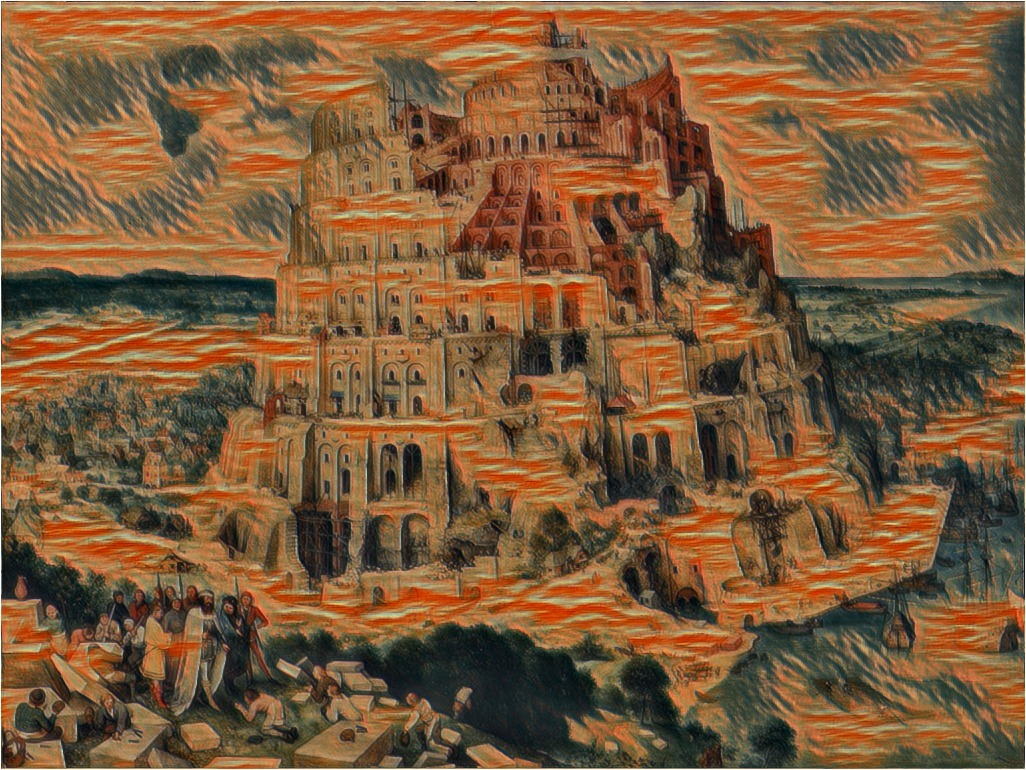}
\caption{Being fully convolutional, our architecture is not bound to image resolution it was trained with. Above, a network trained to stylize 256$\times$256 images was applied to 1024$\times$768 reproduction of Pieter Bruegel's ``The Tower of Babel''.}
\end{figure*}

\begin{figure*}
\centering
\includegraphics[width=\textwidth]{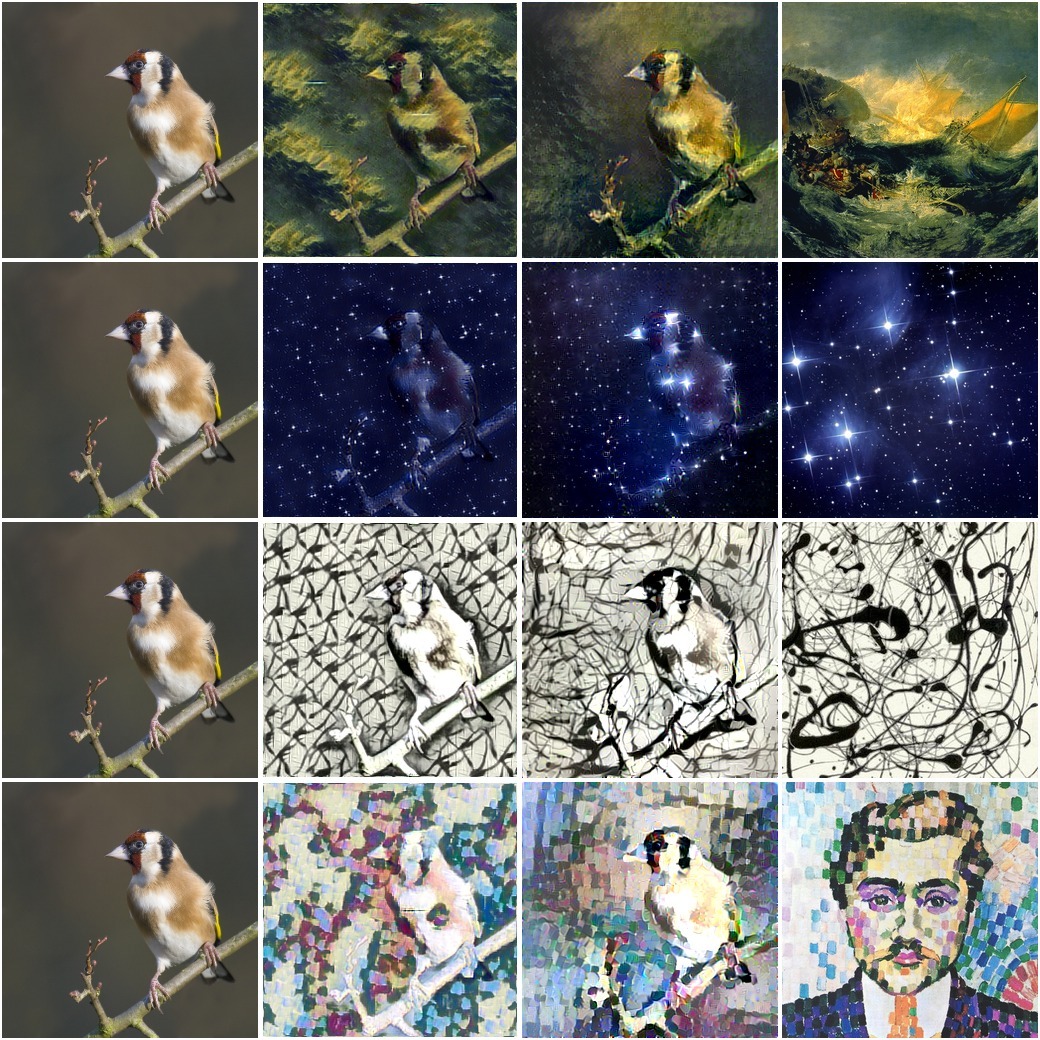}\\
\makebox[0.25\textwidth]{Content}%
\makebox[0.25\textwidth]{Texture nets (ours)}%
\makebox[0.25\textwidth]{Gatys~et~al.}%
\makebox[0.25\textwidth]{Style}\\
\caption{More style transfer comparisons with Gatys~et~al. For the styles above, the results of our approach are inferior to Gatys~et~al. It remains to be seen if more complex losses or deeper networks can narrow down or bridge this performance gap.}
\end{figure*}

\end{document}